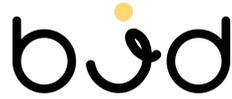

# Efficient Hybrid Decoding for LLMs: Reward-Based Token Modelling with Selective Cloud Assistance

Adarsh M S, Jithin V G, Ditto P S

Bud Ecosystem Inc

## Abstract

Large language models (LLMs) are known for their exceptional performance across a range of natural language processing tasks, but their deployment comes at a high computational and financial cost. On the other hand, smaller language models (SLMs), which can be deployed on lower-cost edge devices, struggle to match the performance of their larger counterparts. This paper presents a novel hybrid inference approach that leverages the strengths of both model types while minimizing reliance on costly cloud-based LLMs. Unlike existing methods that route entire queries to either an SLM or a cloud LLM, our approach introduces a reward-based mechanism to dynamically determine the involvement of the cloud LLM during token generation. Specifically, each token predicted by the SLM is evaluated against a reward score, and only when this score falls below a certain threshold is the cloud LLM consulted for assistance in the next token prediction. This method not only reduces the traffic to the cloud LLM, thereby lowering costs, but also allows for flexible control over response quality depending on the reward score threshold. Experimental results demonstrate that our approach significantly reduces cloud LLM usage with minimal impact on overall response quality, offering a cost-effective solution for deploying high-performance language models.

## 1. Introduction

Large language models (LLMs) have quickly become the gold standard in natural language processing, excelling in tasks like text generation, translation, and code completion. Their impressive performance is largely due to their vast scale, with many models containing billions of parameters. However, the computational and memory requirements of these models are substantial, often necessitating deployment on expensive cloud infrastructure, which drives up operational costs.

To mitigate these costs, there is a growing focus on developing smaller language models (SLMs) that can be deployed on more affordable, edge-based devices. While these models are more cost-efficient, they typically struggle to match the response quality of their larger counterparts. Existing hybrid inference or routing approaches, such as those proposed by (Ding et al., n.d.; Ong et al., 2024), attempt to bridge this gap by routing entire queries to either an SLM or an LLM based on predicted query difficulty. Although effective in reducing costs, traditional binary routing strategies, where a query is handled entirely by either the SLM or the cloud LLM, are effective in reducing costs but may lead to suboptimal performance, especially with more complex queries. To address this limitation, we enhance the routing process by evaluating each token individually, enabling a more refined collaboration between the SLM and cloud LLM. This ensures that cloud resources are only utilized when necessary, preserving both efficiency and response quality.

Our novel hybrid inference approach introduces reward-based token modeling, leveraging reward models typically used in Proximal Policy Optimization (PPO) for model alignment (Zheng et al., 2023). In this method, each token generated by the SLM is evaluated based on a reward function, which measures its alignment with the cloud LLM's probability distribution. Tokens scoring above a set threshold are accepted, while those scoring below are discarded, prompting the cloud LLM to generate the next token. This ensures the SLM receives assistance only when required, optimizing the balance between high-quality outputs and reduced reliance on cloud resources.

The primary contributions of our work are as follows: (1) We introduce a reward-based modelling approach that provides fine-grained control over cloud LLM involvement by dynamically assessing token alignment in real-time. (2) We implement a selective assistance mechanism where the cloud LLM takes over token generation only when the SLM's output falls below a certain quality threshold, effectively reducing computational overhead. (3) We provide extensive experimental validation of our approach, demonstrating its effectiveness in reducing cloud usage with minimal impact on overall performance.

In an era where the demand for high-quality natural language generation continues to grow, our approach offers a flexible and cost-effective solution, empowering both developers and consumers to harness the capabilities of LLMs without incurring prohibitive costs.

## 2. Methodology

In this section, we outline our proposed hybrid inference approach designed to optimize the collaboration between small language models (SLMs) and large language models (LLMs). Our methodology leverages a reward-based token modelling framework to dynamically assess the quality of tokens generated by the SLM. Based on this assessment, we selectively invoke cloud LLM when necessary, ensuring a balance between maintaining high response quality and minimizing computational costs. The following subsections detail the key components of our approach, including the reward function, selective assistance mechanism, and the decoding architecture that integrates these elements.

## 2.1. Reward Model

To enable per-token routing decisions, our approach requires a scoring function that can generate precise token-level scores. Unlike the query-based decision-making in prior methods like HybridLLMs ([Ding et al., n.d.](#)), we utilize a reward model inspired by Reinforcement Learning with Human Feedback (RLHF) ([Zheng et al., 2023](#); [Huang et al., 2024](#)). This model assigns scores to each token, providing the fine-grained control needed to manage the interaction between small language models (SLMs) and large language models (LLMs).

Our reward model is derived from pre-trained transformer-based language models. We adapt these models by removing the final unembedding layer and introducing an additional linear layer to the last transformer layer. ([Huang et al., 2024](#))

### 2.1.1 Choosing the Reward Model

The selection of an appropriate reward model is critical and should be guided by the characteristics of both the SLM and the cloud LLM. The effectiveness of routing depends on how closely the SLM's output aligns with the cloud LLM's distribution. Therefore, it is advantageous to select a reward model from the same family as the cloud LLM, as this will facilitate quicker convergence toward the desired distribution.

While a lightweight reward model is crucial for minimizing inference latency and computational overhead, a balance between size and capability is necessary. A smaller model helps maintain efficiency in per-token routing decisions. However, if the model is too small, it may not accurately capture the complex probability distribution of the larger LLM, leading to suboptimal routing decisions. This misalignment can result in frequent, unnecessary calls to the LLM or failure to recognize when LLM assistance is needed. Thus, the reward model must be both compact and sufficiently sophisticated to effectively align with the cloud LLM's distribution.

### 2.1.2 Data Synthesis and Length Bias Mitigation

The quality and diversity of the training data are critical to the effectiveness of our reward model. In traditional Reinforcement Learning with Human Feedback (RLHF) approaches, reward models are trained using paired comparisons between responses, typically annotated with human preferences ([Zheng et al., 2023](#)). However, since our approach focuses on aligning the reward model with the cloud LLM's probability distribution rather than human preferences, such data is less relevant. Instead, we require synthetic datasets that reflect the token-level alignment between the chosen (cloud LLM) and rejected (SLM) distributions, rather than approximating human-desired output quality.

To address this, synthetic data generation methods become crucial. [Bai et al. (2022b)](#) propose Reinforcement Learning from AI Feedback (RLAIF), where large language models are used to label response pairs, bypassing human annotation. Similarly, [Kim et al., 2023](#) suggest generating positive and negative responses using models of different quality, which is conceptually similar to our approach. We also synthesize data based on different model distributions, allowing our reward model to effectively judge token alignment and ensure more accurate routing decisions. These methods offer a scalable way to produce training data that matches the cloud LLM's distribution, optimizing the reward model's performance.

To meet these requirements, we propose the following data synthesis procedure:

**1. Model serving**

Both the SLM and cloud LLM are run in a high-throughput serving engine, such as vLLM ([Vllm-Project, n.d.](#)). This setup allows us to efficiently generate responses by accessing the completion endpoints of both models. The generated responses serve as the foundation for the paired comparison dataset, where responses from the cloud LLM represent the chosen (preferred) distribution, and responses from the SLM represent the rejected (dispreferred) distribution.

**2. Prompt Space Creation**

The prompt space from which responses are generated must be diverse and representative to ensure the reward model generalizes well. We achieve this by sampling prompts from one or more large language model (LLM) datasets. It is advisable to sample from multiple datasets to create a sparse and varied prompt space, enhancing the reward model's ability to understand and align with the probability distribution of the cloud LLM.

**3. Response Generation**

For each prompt, responses are generated from both the SLM and the cloud LLM. Instead of using fixed-length sequences, the responses are chunked into segments where the chunk length varies from 1 to the minimum length of the SLM and LLM responses. This approach ensures that the reward model is exposed to both shorter and longer sequences during training. Importantly, for each response pair, multiple chunks can be generated, but within each chunk, the SLM and LLM responses will have the same token length to maintain consistency in token-level comparison.

1. **Reducing Length Bias:** By including chunks as short as a single token, this method addresses the tendency of reward models to favour longer responses, as observed in existing studies ([Singhal et al., 2023](#)). The inclusion of single-token chunks enables the model to learn fine-grained token-level differences, reducing the inherent bias towards longer sequences.
2. **Balancing Token-Level and Sequence-Level Evaluation:** The variable chunk length allows the model to evaluate both short and long segments of text, providing a more balanced training dataset. This ensures that the reward model can assess responses of varying lengths, from individual tokens to nearly complete sequences, fostering a more nuanced understanding of both token-level and sequence-level interactions.

To enhance the dataset's distribution coverage and diversity, it is recommended to perform response generation in sampling mode, with randomized *temperature* and *top-p* parameters for each generation. This randomization encourages exploration of a wide range of potential token sequences, resulting in a more comprehensive dataset and, consequently, a more effective reward model.

**Length Bias in Reward Models**

Addressing length bias is crucial because reward models in RLHF method often exhibit a preference for longer responses. This bias, as shown in previous research ([Singhal et al., 2023](#)), can skew the model's scoring function, making it less sensitive to shorter, more concise responses. In their work, a balancing strategy was used to subsample examples based on length differences in order to create a more symmetric distribution of long and short responses. In contrast, our approach leverages a chunking strategy that spans a range of lengths, which exposes the reward model to varying lengths during training, rather than merely balancing the dataset. This allows the reward model to remain versatile and robust, accurately scoring both shorter and longer segments, ultimately leading to more balanced and effective per-token routing decisions.

### 2.1.3 Training

Once the dataset has been synthesized, the training process for the reward model can proceed using the conventional approach employed in RLHF (Zheng et al., 2023; Huang et al., 2024). The dataset consists of paired comparisons between responses, with each pair containing a chosen (cloud LLM) and a rejected (SLM) response. The reward model is trained to score tokens such that higher scores indicate alignment with the chosen distribution, while lower scores reflect alignment with the rejected distribution.

The loss function for training the reward model is defined as:

$$L(\psi) = log\sigma(r(x, yw) - r(x, yl))$$

Here, $r(x, yw)$ and $r(x, yl)$ represent the reward scores for the preferred (chosen) and dispreferred (rejected) responses, respectively. This loss function drives the model to correctly differentiate between the distributions represented by the cloud LLM and the SLM, ultimately enabling it to make accurate token-level routing decisions during inference.

A lower preference loss is desired, as it indicates better judgement by the reward model in predicting how well a token aligns with the cloud LLM's probability distribution. Achieving a lower preference loss reflects improved alignment with the cloud LLM's distribution and more accurate token-level routing decisions during inference. This, in turn, enhances the model's ability to effectively utilize the cloud LLM's capabilities only when necessary, while maximizing the efficiency of the SLM.

## 2.2. Selective Assistance

In conventional Proximal Policy Optimization (PPO) architectures used in RLHF, although per-token scores are generated by the reward model, typically only the score associated with the End-of-Sequence (EOS) token is utilized as the final reward (Huang et al., 2024). While this sequence-level reward extraction has its applications, it is less suited to architectures like ours that require token-level decision-making throughout the generation process. Therefore, our approach focuses on per-token rewards to optimize for selective token routing, rather than waiting until the sequence's end for evaluation.

Recent work by Yang et al. (2023) addresses a related challenge in aligning token-level training with sequence-level preferences, proposing a method to ground sequence-level preferences into token-level training guidance. Although we do not directly adopt their approach, this concept offers potential improvements for future iterations of our method by further enhancing token-level alignment without compromising sequence coherence. Exploring such techniques could further strengthen the reward model's capacity to guide token-level decisions more precisely.

In this paper, we extend the speculative decoding architecture introduced by Leviathan et al. (2022) to improve the token generation process by incorporating a candidate generator model. In speculative decoding, a candidate generator (usually a smaller model) proposes candidate tokens at each decoding step. The large model then evaluates these candidates and generates the next token based on the current prefix. If a candidate token is accepted by the large model, it is concatenated with the next token generated by the large model and added to the growing sequence. If the candidate is rejected, only the token generated by the large model is used, replacing the candidate token. This iterative process continues until the full sequence is generated.

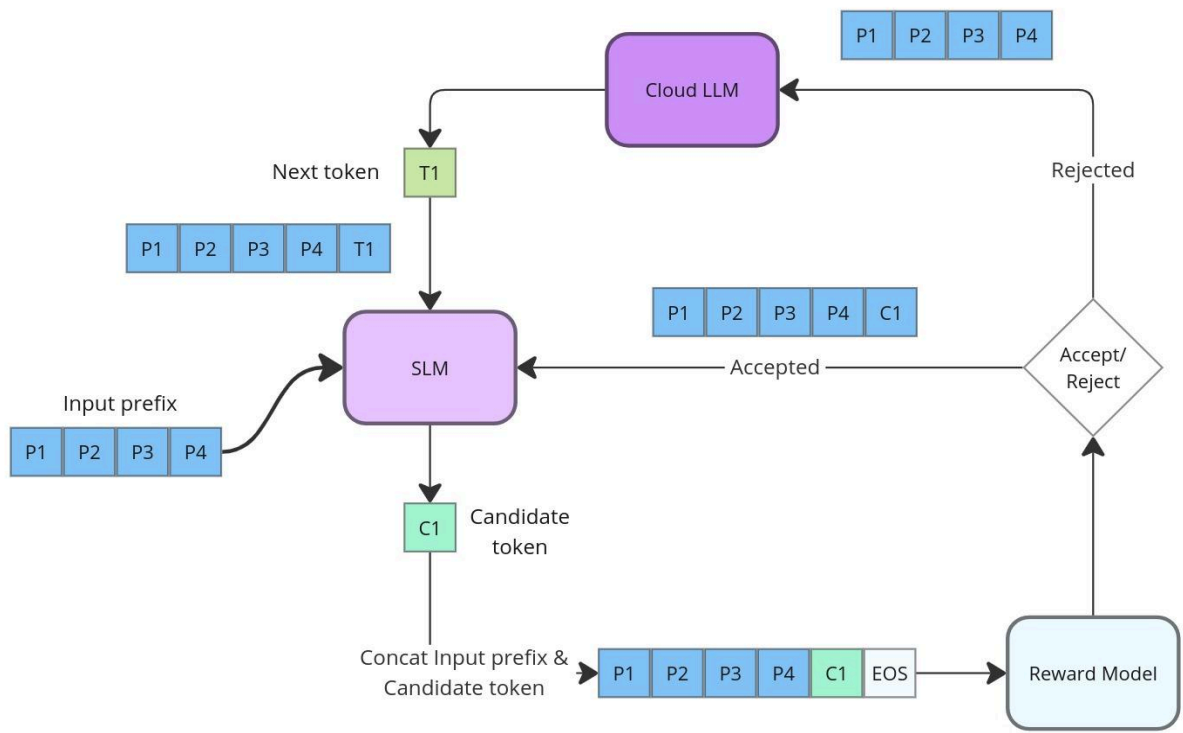

*Figure 1*: This figure illustrates the selective assistance process, where the SLM generates candidate tokens that are scored by the reward model. If a token's score exceeds the threshold, it is accepted and added to the sequence. If rejected, the cloud LLM generates the next token. This process continues iteratively until the sequence is complete, optimizing response quality by selectively involving the cloud LLM.

Our approach adapts the speculative decoding framework by leveraging the SLM as the candidate generator model. In each decoding step, the SLM generates candidate tokens, and our reward model evaluates each token, assigning a score based on how well it aligns with the desired distribution (typically that of the cloud LLM). The key innovation in our approach is the application of a threshold-based acceptance criterion:

- **Acceptance:** If the reward score of the candidate token exceeds a predefined threshold, the token is accepted, and the decoding flow continues with the SLM generating the next candidate token.

- **Rejection:** If the reward score falls below the threshold, the candidate token is rejected, and the prefix used to generate this token is sent to the cloud LLM. The cloud LLM then generates the next token.

This selective assistance mechanism ensures that the cloud LLM is engaged only when the SLM's token generation falls short of the desired quality, as determined by the reward model. By doing so, the system balances the computational efficiency of the SLM with the higher-quality output of the cloud LLM, effectively reducing reliance on the cloud LLM while maintaining high overall output quality.

This approach also allows for fine-grained control over the generation process, enabling dynamic adjustment of the threshold based on context or performance needs. For instance, in scenarios where higher precision is required, the threshold can be raised, increasing the likelihood that the cloud LLM will be invoked. Conversely, in less critical contexts, the threshold can be lowered, allowing the SLM to handle more of the generation, thus reducing inference latency and cost.

By employing token-level selective assistance, our architecture ensures that each token is evaluated on its own merit, enabling a more precise and efficient allocation of computational resources between the SLM and the cloud LLM. This method provides a robust framework for deploying hybrid models in real-world applications, where balancing quality and efficiency is paramount.

## 3. Experiments

We conducted experiments using the Qwen2 family of models, selected for their range of parameter variants, to validate our proposed hybrid decoding approach. Specifically, we utilized the Qwen 2-0.5B-Instruct model as the reward model due to its efficiency as the smallest model in the series. For data synthesis, the Qwen2-1.5B-Instruct model was employed as the smaller language model (SLM), while the Qwen2-7B-Instruct model served as the larger language model (LLM).

**Dataset Creation**

The prompt space for the dataset was constructed by sampling from a diverse set of RLHF-related datasets, including *Anthropic/hh-rlhf*, *OpenAI/webgpt_comparisons*, *Tasksource/oasst1_pairwise_rlhf_reward*, and *LMSys/chatbot_arena_conversations*. The initial prompt space was randomly sampled to create a dataset of 100k samples. After applying our chunking method, the synthesized dataset expanded to 300k samples. From this augmented dataset, we randomly selected 100k samples for training the reward model.

**Training and Evaluation**

The reward model training yielded a low preference loss, indicating strong alignment with the LLM's probability distribution and effective differentiation between tokens that align with this distribution and those that do not. The training was conducted on two 80GB A100 GPUs, using a batch size of 256, which allowed for efficient processing of the large dataset.

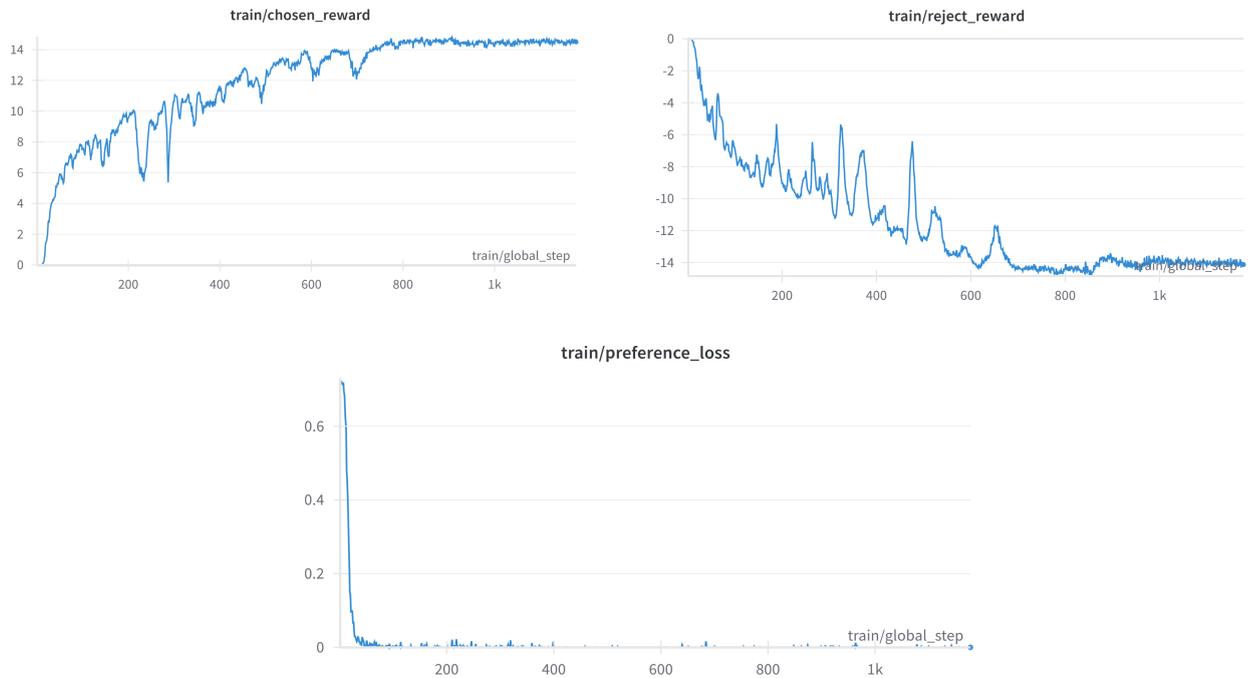

*Figure 2*: The three figures above collectively illustrate the effectiveness of the reward model in distinguishing between chosen and rejected tokens:

- *Train/Chosen Reward*: Displays the reward scores assigned to tokens generated by the LLM, highlighting their alignment with the probability distribution.
- *Train/Reject Reward*: Represents the reward scores for tokens produced by the SLM that were rejected, showing the model's capacity to identify misaligned tokens.
- *Train/Preference Loss*: Demonstrates the overall preference loss during training. The consistently low loss signifies the model's effectiveness in maintaining a clear distinction between chosen and rejected tokens.

To evaluate the efficacy of our hybrid decoding, we performed various evaluations on the architecture. The results, presented in the following tables, demonstrate the potential of our approach to balance quality and efficiency in token generation by selectively involving the cloud LLM only when necessary.

| Dataset | Reward threshold | SLM accuracy (Baseline) | Hybrid decoding accuracy | LLM accuracy (Baseline) | Cloud LLM activation ratio |
| --- | --- | --- | --- | --- | --- |

| | | | (Ours) | | (%) |
|---|---|---|---|---|---|
| gsm8k | 1.0 | 50.7 | 66.48 | 77.8 | 56.00 |
| | 2.0 | | 74.75 | | 78.00 |
| | 4.0 | | 77.78 | | 87.00 |
| mmlu | 1.0 | 52.4 | 69.9 | 70.5 | 92.00 |
| mbpp | 1.0 | 36.6 | 38.8 | 60.00 | 35.00 |
| | 1.5 | | 52.2 | | 68.00 |
| | 2.0 | | 60.0 | | 87.00 |
| | 4.0 | | 60.0 | | 98.00 |
| cnndm | 1.0 | 20.9 | 25.7 | 28.9 | 59.00 |

*Table 1*: The table presents the evaluation metrics across different datasets. It includes the dataset name, the reward threshold used for evaluation, and the accuracy of both the SLM (Small Language Model) and the LLM (Large Language Model), which serve as baselines. Additionally, it displays the accuracy achieved using our proposed hybrid decoding method and the Cloud LLM activation ratio, calculated as the total LLM steps divided by the total SLM steps.

### 3.1.1 Reward Threshold v/s Cloud LLM activation ratio

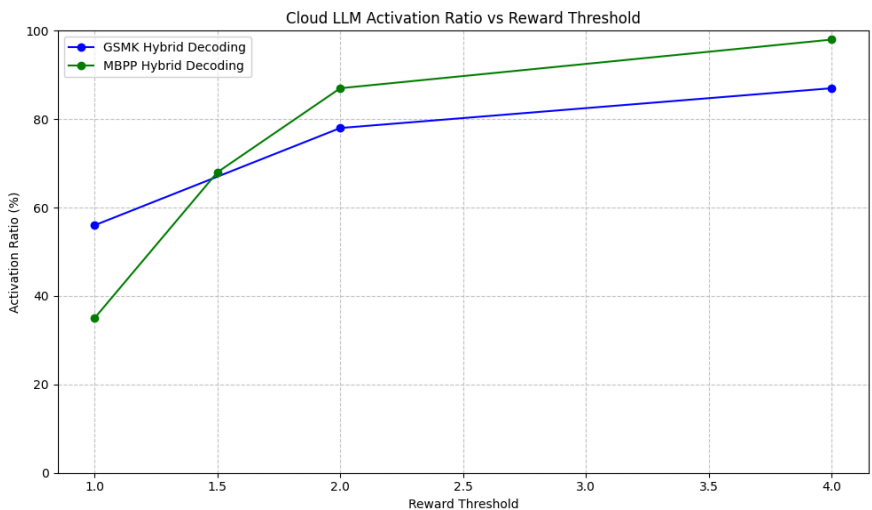

*Figure 3*: Reward threshold v/s Cloud LLM activation ratio relationship for GSM8k and MBPP

The reward threshold plays a pivotal role in determining the activation of the cloud LLM during the hybrid decoding process. A higher threshold indicates stricter criteria for accepting tokens generated by the SLM, leading to increased reliance on the cloud LLM for token predictions. Conversely, a lower threshold allows more tokens from the SLM to be accepted, thereby reducing the activation ratio of the cloud LLM and improving efficiency.

As observed in the evaluation results across various datasets, the cloud LLM activation ratio directly correlates with the reward threshold. For instance, in the GSM8K dataset, setting the reward threshold to 1.0 results in a cloud LLM activation ratio of 56%. As the threshold increases to 2.0 and 4.0, the activation ratio rises to 78% and 87%, respectively. A similar trend is seen in the MBPP dataset, where increasing the threshold from 1.0 to 4.0 leads to a significant rise in cloud LLM activation, from 35% to 98%. This pattern holds across most datasets, demonstrating that a higher reward threshold drives the architecture towards heavier reliance on the cloud LLM for generating tokens, particularly in more complex or nuanced tasks.

Interestingly, while a higher threshold improves hybrid decoding accuracy, it comes at the cost of increased cloud LLM usage, which reduces the efficiency gains intended by the hybrid mechanism. Balancing the threshold is crucial to maintain the optimal trade-off between leveraging the SLM for routine tokens and activating the cloud LLM only when necessary for more complex or ambiguous token predictions.

### 3.1.2 Reward Threshold v/s Accuracy

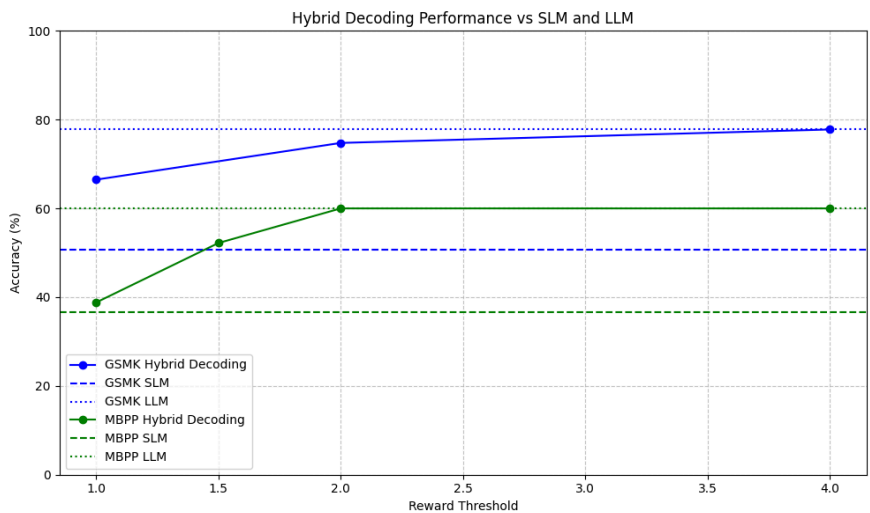

*Figure 4*: Reward threshold v/s Accuracy relationship for GSM8k and MBPP compared to the SLM and LLM baseline.

The reward threshold significantly influences the accuracy of our hybrid decoding system as seen in the evaluation results. By adjusting the threshold, we control the strictness with which the reward model accepts tokens generated by the SLM. A lower threshold allows more SLM-generated tokens to be accepted, which may improve efficiency but can result in lower overall accuracy. Conversely, a higher threshold results in more frequent activation of the cloud LLM, improving accuracy but at the cost of increased resource usage.

For the GSM8K dataset, a reward threshold of 1.0 achieves 66.48% accuracy in hybrid decoding, significantly higher than the SLM baseline of 50.7% but lower than the LLM's 77.8%. As the threshold increases to 2.0, hybrid accuracy improves to 74.75%, closely matching the cloud LLM's performance. At a threshold of 4.0, hybrid accuracy peaks at 77.78%, closely approximating the LLM's baseline accuracy. This trend highlights how increasing the threshold allows the hybrid model to approach the performance of the LLM, particularly for more challenging tasks where SLM performance might lag.

A similar pattern is evident in the MBPP dataset, where increasing the reward threshold from 1.0 to 2.0 raises hybrid accuracy from 38.8% to 60.0%, approaching the LLM's 60% accuracy. For more challenging datasets like MMLU, the trend is slightly different; the cloud LLM activation occurs at a higher rate (92%) even with a reward threshold of 1.0, leading to a hybrid accuracy of 69.9%, close to the LLM's 70.5% accuracy.

Overall, a higher reward threshold results in better accuracy because the reward model is more likely to reject SLM tokens that deviate from the LLM's probability distribution. However, this comes at the cost of efficiency, as more tokens will require cloud LLM intervention. The challenge lies in selecting an optimal threshold that balances the accuracy gains of the hybrid system with the need for efficient computation, particularly in scenarios where computational cost is a limiting factor.

| Dataset | Reward threshold | SLM throughput (Baseline) | Hybrid decoding throughput (Ours) | LLM throughput (Baseline) | Cloud LLM activation ratio (%) |
|---|---|---|---|---|---|
| gsm8k | 1.0 | 35.18 | 10.70 | 34.05 | 56.00 |
| gsm8k | 2.0 | 35.18 | 8.71 | 34.05 | 78.00 |
| gsm8k | 4.0 | 35.18 | 8.48 | 34.05 | 87.00 |
| mbpp | 1.0 | 22.38 | 6.39 | 18.62 | 35.00 |
| mbpp | 1.5 | 22.38 | 4.82 | 18.62 | 68.00 |
| mbpp | 2.0 | 22.38 | 4.20 | 18.62 | 87.00 |
| mbpp | 4.0 | 22.38 | 4.10 | 18.62 | 98.00 |
| cnndm | 1.0 | 23.71 | 4.77 | 19.22 | 59.00 |

*Table 2*: The table presents the throughput evaluation across different datasets. It includes the dataset name, the reward threshold used for evaluation, and the throughput of both the SLM (Small Language Model) and the LLM (Large Language Model), which serve as baselines. Additionally, it displays the throughput achieved using our proposed hybrid decoding method and the Cloud LLM activation ratio, calculated as the total LLM steps divided by the total SLM steps. The table highlights how changes in the reward threshold affect both the throughput and cloud LLM utilization.

### 3.1.3 Throughput & Latency

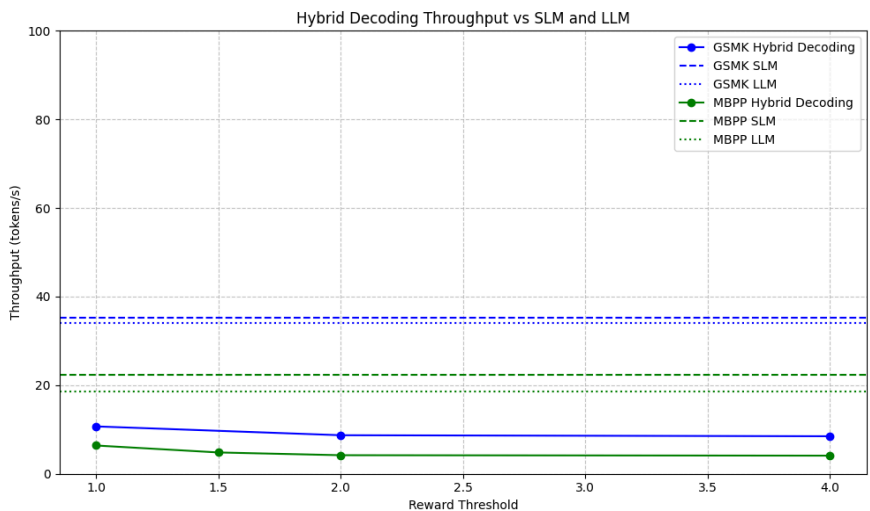

*Figure 5*: Reward threshold v/s Throughput relationship for GSM8k and MBPP compared to the SLM and LLM baseline.

The throughput analysis in *Table 2* highlights the trade-offs between the reward threshold and the overall efficiency of our hybrid decoding framework. As the reward threshold increases, the throughput of the hybrid decoding method decreases across all datasets, with a corresponding increase in cloud LLM activation. This decrease in throughput is primarily due to the increased reliance on the cloud LLM, which has a higher computational cost and longer latency per token generation.

**Impact of Reward Threshold on Throughput**: At lower reward thresholds (e.g., 1.0), the hybrid decoding throughput is significantly higher compared to higher thresholds, especially for datasets like GSM8K and MBPP. This is because more tokens are accepted from the SLM, allowing the faster small model to contribute more to the decoding process. However, as the reward threshold increases, the cloud LLM is invoked more frequently, leading to a drop in throughput. For instance, in the MBPP dataset, the throughput drops from 6.39 to 4.10 as the reward threshold moves from 1.0 to 4.0, with a corresponding spike in cloud LLM activation from 35% to 98%.

**Trade-off Between Throughput and Quality**: The observed drop in throughput as the reward threshold increases is expected since the hybrid framework prioritizes the generation quality by relying more on the cloud LLM. As the threshold rises, fewer tokens are accepted from the SLM, increasing the dependence on the cloud LLM's more accurate but slower token generation, which lowers overall throughput.

**Latency Considerations**: The latency of the hybrid decoding method is closely tied to cloud LLM activation. When the cloud LLM is activated frequently (as seen with higher reward thresholds), latency increases because of the larger model's computational cost. This is evident in datasets like CNN/DM, where the activation ratio jumps to 59% at a threshold of 1.0, resulting in decreased throughput of 4.77 compared to the SLM baseline throughput of 23.71.

**Balancing Efficiency and Performance**: For applications that prioritize efficiency, a lower reward threshold can maintain higher throughput with minimal reliance on the cloud LLM. However, for applications where accuracy is paramount, the trade-off of reduced throughput in favor of higher reward thresholds and cloud LLM activation may be acceptable. This balance between throughput and quality is particularly evident in high-complexity datasets like MBPP, where the throughput of the hybrid decoding method remains competitive even at higher reward thresholds.

## 4. Limitations

While our proposed architecture offers significant advantages in terms of efficiency and flexibility, several limitations should be noted:

**Reward Model Alignment & Routing Efficiency**: The efficiency of the routing mechanism depends heavily on the reward model's ability to accurately assess token alignment with the cloud LLM's probability distribution. Flaws in the reward model's judgment can lead to either over-reliance on the SLM, reducing output quality, or excessive use of the cloud LLM, diminishing efficiency. Moreover, the binary nature of the routing decision—accepting or rejecting tokens—requires recalibration whenever a new SLM-LLM pair is introduced. This necessitates retraining and data augmentation, which can be resource-intensive, especially in environments with frequent model updates.

**Inference Time Overhead**: Incorporating the reward model into the inference pipeline can introduce additional latency, particularly in scenarios where a large number of tokens are rejected. In the worst-case scenario, where every candidate token generated by the SLM is rejected, the system must rely on the cloud LLM for every token, leading to a significant increase in total inference time. This overhead can be especially pronounced in applications requiring real-time or low-latency responses.

**Scalability Concerns**: As the number of potential SLM and cloud LLM pairs grows, the complexity of managing and maintaining reward models for each pair increases. The system must ensure that the reward model for each pair remains well-aligned, which may not scale efficiently in practice, especially in diverse or dynamic deployment environments.

**Dependency on Threshold Tuning**: The effectiveness of the token selection process relies on the appropriate tuning of the acceptance threshold. If the threshold is set too high, the system may reject many valid tokens from the SLM, unnecessarily increasing reliance on the cloud LLM. Conversely, if set too low, the system may accept suboptimal tokens, degrading the overall quality of the output. Finding and maintaining this balance is challenging and may require ongoing adjustments based on the specific application or evolving model behaviours.

## 5. Future Considerations

**Implementing a Cache Mechanism**: Inspired by the Ouroboros approach ([Zhao et al., 2024](#)), a cache mechanism can be implemented to store previously accepted tokens generated by the LLM. This would reduce redundant calls to the LLM by reusing tokens when similar contexts arise. By minimizing the number of calls to the cloud LLM, we can significantly reduce overall latency, especially during repetitive or predictable token generation scenarios. This cached approach can be further enhanced with a lookahead decoding strategy ([Fu et al., 2024](#)), where multiple tokens are predicted ahead of time, validated, and cached, streamlining the inference process.

**Introducing a KV Cache for LLM**: The current implementation does not utilize a key-value (KV) cache for the LLM, which could otherwise help reduce inference time. By caching attention keys and values from previous decoding steps, the model could skip redundant computations, leading to faster token generation for large sequences. Incorporating a KV cache would not only enhance performance but also reduce the overhead associated with the cloud LLM's repeated computations during decoding.

**Improving the Reward Model**: While the current reward model performs well, there is potential for enhancement in its judgment of token alignment with the LLM's probability distribution. This can be achieved through more sophisticated training techniques, better optimization, or incorporating additional features such as fine-tuning on more diverse datasets. [Yang et al. (2023)](#) address the challenge of sequence-level preferences versus token-level training by grounding sequence-level preferences into token-level guidance. Adapting similar principles could further refine our approach, leading to more accurate routing decisions and fewer LLM calls, thus improving overall efficiency.

**Dynamic Reward Threshold Adjustment**: Another potential improvement is implementing dynamic reward thresholding. Instead of using a static threshold throughout the decoding process, the threshold could adjust dynamically based on factors such as context complexity or model confidence. This would allow for more flexible and efficient routing, adapting to the specific requirements of each token being processed. For simpler tokens, the SLM could handle more of the load, while complex tokens could be routed to the cloud LLM only when absolutely necessary.

**SLM-LLM Alignment for Enhanced Efficiency:** Aligning the SLM more closely with the LLM can improve the efficiency of the system by reducing the disparity between the two models. Research such as [Shen et al. (2023)](#) and [Goel et al. (2024b)](#) suggests that better alignment can increase the number of accepted tokens in speculative decoding scenarios. By aligning the SLM more effectively with the LLM, we can enhance token generation accuracy and reduce the need for LLM assistance, ultimately improving system performance.

**Exploring Multi-Model Collaboration**: Future work could explore multi-model collaboration, where different types of models (e.g., SLMs specialized in specific tasks) collaborate with the cloud LLM. This would allow for a more modular approach to token generation, where each model contributes based on its strengths. This could further optimize both performance and accuracy, enabling selective assistance not just between two models but across a network of specialized models.

By exploring these avenues, we can further improve the efficiency, accuracy, and scalability of the hybrid decoding framework, making it more suitable for real-world deployment in latency-sensitive applications.

## Conclusion

In this paper, we introduced a novel hybrid decoding framework that leverages the strengths of both small language models (SLMs) and large language models (LLMs) to achieve efficient and high-quality token generation. By employing a reward model to make per-token routing decisions, our approach balances the trade-off between computational efficiency and output accuracy. The experiments demonstrated that our hybrid decoding method significantly reduces the reliance on the cloud LLM, while maintaining competitive accuracy across various tasks.

We addressed critical challenges such as length bias in reward models and explored strategies to mitigate it through flexible chunking mechanisms. Furthermore, we showed that the reward threshold directly influences the accuracy, throughput, and cloud LLM activation ratio, highlighting the importance of fine-tuning this parameter for different use cases.

Our work opens up new possibilities for more efficient language model inference, especially in scenarios where computational resources are limited or cost-sensitive. Future improvements, such as implementing caching mechanisms, KV cache utilization, and reward model refinement, will further enhance the performance and scalability of the system, making it even more robust and applicable to a wide range of real-world applications.